# A Generalized Flow for Multi-class and Binary Classification Tasks: An Azure ML Approach

Matthew Bihis
Student, Electrical Engineering
University of Washington
Bothell, USA
bihism@uw.edu

Sohini Roychowdhury
Assistant Professor, Electrical Engineering
University of Washington
Bothell, USA
roych@uw.edu

*Abstract:* The constant growth in the present day real-world databases pose computational challenges for a single computer. Cloud-based platforms, on the other hand, are capable of handling large volumes of information manipulation tasks, thereby necessitating their use for large real-world data set computations. This work focuses on creating a novel Generalized Flow within the cloud-based computing platform: Microsoft Azure Machine Learning Studio (MAMLS) that accepts multi-class and binary classification data sets alike and processes them to maximize the overall classification accuracy. First, each data set is split into training and testing data sets, respectively. Then, linear and non-linear classification model parameters are estimated using the training data set. Data dimensionality reduction is then performed to maximize classification accuracy. For multi-class data sets, data-centric information is used to further improve overall classification accuracy by reducing the multi-class classification to a series of hierarchical binary classification tasks. Finally, the performance of optimized classification model thus achieved is evaluated and scored on the testing data set. The classification characteristics of the proposed flow are comparatively evaluated on 3 public data sets and a local data set with respect to existing state-of-the-art methods. On the 3 public data sets, the proposed flow achieves 78-97.5% classification accuracy. Also, the local data set, created using the information regarding presence of Diabetic Retinopathy lesions in fundus images, results in 85.3-95.7% average classification accuracy, which is higher than the existing methods. Thus, the proposed generalized flow can be useful for a wide range of application-oriented "big data sets".

*Keywords—Microsoft Azure Machine Learning Studio; Generalized Flow; Big Data; Classification; Fundus Images*

## I. INTRODUCTION

Recent years have seen a huge emergence of Big Data analytics in various real-life applications such as Business Optimization, Health Care Research, Financial Trading, Data Visualization, Medical/Healthcare Data, Internet of Things (IoT), to name a few [1]. Processing Big Data typically requires efficient management of the following features: volume, velocity, variety, and complexity [1]. Numerous approaches have been proposed on the basis of advances, storage, and computation paradigms. These can be broadly categorized as:
- Single-node in-memory analytics with multi-threaded computation over data cached in-memory. For example: MATLAB, R, Weka, RapidMiner, etc.
- Distributed Storage with Parallel/Distributed Computing where data is stored over multiple clusters, as in Hadoop Distributed File System, and computation done in parallel, through Message Passing Interface (MPI).

Most Big Data real-life applications involve adapting these advanced paradigms for Machine Learning (ML) algorithms. The Microsoft Azure Machine Learning Studio (MAMLS) is a cloud-based computing platform that is accessible through a web-based interface. A variety of ML algorithms are available through classification model modules developed by Microsoft. In this work, the existing ML modules in MAMLS are modified and combined with modules in 'R' language to achieve a generalized flow that analyzes a small portion of a "big data" set, which can be acquired from a varied range of applications, and maximizes multi-class and binary classification accuracies with minimal manual intervention.

A wide range of classifiers have been analyzed so far for credit score reporting, pathology and anomaly classification problems. For instance, to build credit scoring models, optimized model parameterization followed by feature scoring has been implemented in [3]. Some of the well-known classifiers include logistic regression, decision trees, k-nearest neighbors, linear discriminant analysis, neural networks and support vector machines (SVM) [5]. For pathology classification applications, typically breast cancer detection and diagnostics several neural network based classification models have been implemented to design expert diagnostic systems [8]. Additionally, feature ranking strategies based on Spearman's ranking coefficient and Fisher-scoring metrics have been used extensively for data dimensionality reduction [3] [5] [8]. This paper makes two key contributions. First, we apply several feature ranking strategies and observe that for an optimal set of top ranked features, classification accuracy is maximized for public and local data sets. Second, we implement a generalized flow that is capable of multi-class and binary classification tasks to maximize overall classification accuracy by dimensionality reduction and optimal classifier parameterization. The novel contribution of this paper is a customized modular representation (Generalized Flow) of the cloud-based MAMLS platform that is benchmarked for classification performances on public data sets and is shown to maximize classification accuracy on a local medical image data set. The proposed generalized flow model in the MAMLS platform is shown in Fig.1.

This paper is organized as follows. In Section II, the proposed generalized flow model and the data sets are explained. In Section III, experimental results of the proposed flow on the data sets are presented. Conclusions and discussion are presented in Section IV.

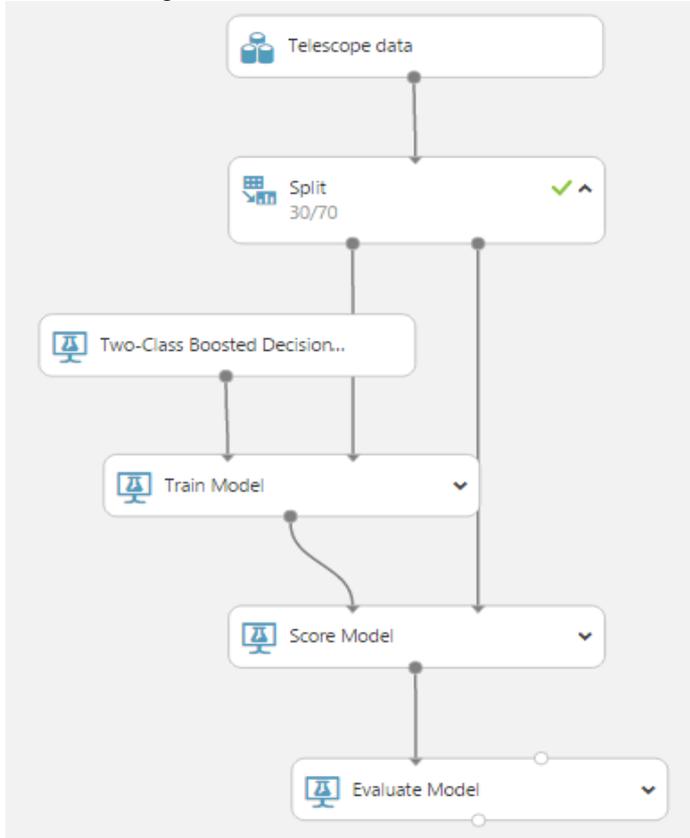

Fig. 1. An example of the MAMLS experiment canvas with Telescope data [6] as input. The data set is split into training (30% of the samples) and testing (70% of the samples) data sets. Next, a Boosted Decision classifier is trained on the training data followed by scoring and evaluating the classifier on testing data set.

## II. DATA, MODELING, AND FEATURE SELECTION

The proposed Generalized Flow in the MAMLS platform requires input data as: $X_{[N \times d]}, Y_{[N \times 1]}$, where 'X' represents 'N' data samples in 'd'-dimensional space, and 'Y' represents the actual class label of each data sample. To maximize the information extracted from each of the data sets, 30% of the samples belonging to each class label become the training data set. The trained classification models are then scored and evaluated on the remaining 70% of the samples in the testing data set. It is noteworthy that 30% of the total number of samples is greater than the dimensionality of the data set, i.e., 0.3N>d, for all the data sets. This method of data partitioning into the training and testing data set is well established in the medical image processing domain [9] and in this work it is extended to other public-domain data sets as well. Thus, classifier parameterization is conducted on the training data set: $X_{[N^{(1)} \times d]}, Y_{[N^{(1)} \times 1]}$, where $N^{(1)} = 0.3N$, and on the testing data set, the probability of each data sample belonging to a particular class 'i' is denoted by $\Pi^i_{[N^{(2)} \times 1]}$, where, $N^{(2)}=0.7N$ (since 0.7N data samples are in the testing data set). For various classification thresholds in the range [0,1], the number of correctly and incorrectly classified samples are then counted to evaluate the precision, recall, accuracy and area under receiver operating characteristic curve (AUC) for classification.

The public and local data set used to evaluate the performance of the proposed generalized flow, the various classifiers under assessment, and dimensionality reduction strategies are given below.

### A. Data

Three public data sets are chosen from the MAMLS database to analyze the performance of the proposed Generalized Flow. Additionally, a local data set is prepared using a retinal (ophthalmic) set of images to assess the performance of an automated cloud-based computing platform on medical image-based pathology detection tasks.

*1) Wisconsin Breast Cancer*
Binary classification of this data set aims to separate normal from pathological patients using ten features and 683 samples [2]. This data set contains 65% samples belonging to class 0 (without cancer) and 35% samples belonging to class 1 (with cancer). An existing state-of-the art method using support vector machines (SVM) for feature based classification achieves an accuracy of 96.6% on this data set [3].

*2) German Credit Card*
Binary classification of this data set aims to predict a customer's risk score based on 20 features and 1000 samples [4]. This data set contains 70% samples belonging to class 0 (low risk) and 30% samples belonging to class 1 (high risk). An existing state-of-the art work using SVM [5] achieves 73.3% classification accuracy on this data. By performing subsequent classification enhancements using Clustering-Launched Classification strategies, the method in [5] achieves 83.3% classification accuracy on this data.

*3) Telescope*
Binary classification of this data set aims to predict gamma bursts from background noise using 10 features and 19020 samples [6]. This data set contains 65% samples belonging to class 0 (gamma; signal) and 35% samples belonging to class 1 (hadron; background).
An existing state-of-the art method using AdaBoost classifier [7] achieves an accuracy of 86.94% on this data set [8].

*4) DIARETDB1*
This data set is locally generated using 89 fundus images acquired with 50° field of view with [1500x1152] pixels each. These fundus images correspond to patients with varying severities of Diabetic Retinopathy (DR). Automated segmentation of these fundus images using the method in [9] leads to extraction of several regions that correspond to different kinds of DR lesions. Region-based and pixel-based features of each lesion region thus extracted are computed and used for pathology classification tasks [9].

The DR lesions can be broadly categorized as bright lesions and red lesions. False positive bright lesions are assigned class label 0, while false positive red lesions have class label 3. The bright lesions, based on their color, texture and shape can be further categorized as hard-exudates (class label 1) and cotton-wool spots (class label 2). Similarly, the red lesions can be categorized as hemorrhages (class label 4) and micro-aneurysms (class label 5). This data set contains 4.9% samples belonging to class 0, 0.18% samples belonging to class 1, 2.6% samples belonging to class 2, 69% samples belonging to class 3, 13% samples belonging to class 4, 10% samples belonging to class 5. Thus, multi-class classification of this data set aims to predict 6 different classes (class label 0 through 5) using 66 features and 15,945 samples [9]. Since the DR lesion classes have a certain hierarchy in their color and texture [9-11], the performance of hierarchical binary classification is compared to multi-class classification. The hierarchies of binary classification are set up as follows:

- Hierarchy level 1: Separation of bright regions from red regions. Classification of regions with class label 0,1,2 from regions with labels 3,4,5.
- Hierarchy level 2: Separation of bright lesion regions from non-lesion regions. Classification of regions with class label 0 from regions with labels 1 and 2. An existing state-of-the art method using Gaussian Mixture Models achieves 89% accuracy in this task [9].
- Hierarchy level 3: Separation of red regions from non-lesion regions. Classification of regions with class label 3 from regions with labels 4 and 5. An existing state-of-the art method using k-Nearest Neighbors achieves 80% accuracy in this task [9].
- Hierarchy level 4: Separation of soft exudates from hard exudates. Classification of regions with class label 1 from regions with class label 2.
- Hierarchy level 5: Separation of micro-aneurisms from hemorrhages. Classification of regions with class label 4 from regions with class label 5.

*B. Modeling*

The MAMLS platform contains a number of classification models that need to be optimally parameterized, and that can be categorized as binary and multi-class classification models. The binary classification models may also be applied to data sets with more than two classes using a one-vs-all approach. All these classifiers are optimally parameterized by minimizing the classification error obtained by 5-fold cross validation of the training data set [7]. The entire training data set is partitioned such that every fifth sample becomes a member of the validation set while training is performed on all remaining samples. The starting point for finding each validation sample is incremented five times such that every sample becomes a part of the validation set at least once as shown in [7]. The following classification models are used to predict $\Pi^i_{[N^{(1)} \times 1]}$, where $i=\{0,1\}$ for binary classification and $i=\{0,1,2,3,4,5,6\}$ for multi-class classification tasks.

*1) Binary Classification Models:*
The following classifiers can be applied for binary decision making tasks:

*a) Support Vector Machine*

This classifier is constructed on the basic principle of estimating a maximum margin separating hyperplane between samples of the two classes in 'd'-dimensional feature space as shown in (1)

$$\min_{(w,\beta,\xi)} \frac{1}{2} w^{(1)T} w^{(1)} + \gamma \frac{1}{2} \sum_{j=1}^{N^{(1)}} \xi^2 \quad (1)$$

such that, $Y_j[w^{(1)T} \varphi(X_j) + \beta^{(1)}] = 1 - \xi_j, j=1,....N^{(1)}$

Here, '$w^{(1)}$' is the weight vector in primal space and '$\gamma$' is the regularizer. For SVM implementation class label Y=0 is changed to Y=-1 and after constructing the Lagrangian and choosing a radial basis kernel function $K(X, X_j)$ that computes the inner product in transformed space, the classification results in (2-3).

$$\Pi = sign(\sum_{j=1}^{N^{(1)}} \alpha^{(1)}_j Y_j K(X, X_j) + \beta^{(1)}) \quad (2)$$

$$given, K(X, X_j) = \varphi(X)^T \varphi(X_j) = \exp(-\gamma \| X - X_j \|^2) \quad (3)$$

*b) Logistic Regression*

The probability of a sample belonging to class $i=1$ is given in (4).

$$\log it(\Pi^1) = \log(\frac{\Pi^1}{1-\Pi^1}) = \alpha^{(2)} + \beta^{(2)T} X \quad (4)$$

Here, $\alpha^{(2)}, \beta^{(2)}$ are the parameters to be estimated.

*c) Boosted Decision Tree*

This category of predictive model maps the feature characteristics to class label values. A decision tree comprises of individual nodes corresponding to the feature values and the leaf nodes correspond to the class labels. Once all the base decision trees are populated using the training data set, each tree in the series is fit to the residuals of the previous tree with the purpose of minimizing classification error. If $T_1,....T_{N^{(1)}}$ are trees, then the model in (5) is realized.

$$F(X) = \beta^{(3)}_0 + \beta^{(3)}_1 T_1(X) + .....\beta^{(3)}_{N^{(1)}} T_{N^{(1)}}(X) \quad (5)$$

Here $\beta^{(3)}_0, \beta^{(3)}_1, .....\beta^{(3)}_{N^{(1)}}$ are coefficients of the boosted tree nodes. Optimal parameterization of this classifier involves optimizing the number of iterations and maximum branch size used in splitting rule [3].

*d) Decision Forest*

Random decision forest classifiers are built from the decision trees, where each tree votes for the most popular class. This tree voting strategy [3] is collectively called random decision trees. The parameters to be tuned include the number of trees and the number of attributes required to grow each tree.

*e) Neural Network*

This category of classifiers is used to model the non-linear association between the features and the class label per sample. A typical neural network comprises of neurons in three layers, the input layer (corresponding to the feature values), a hidden layer, and an output layer (corresponding to the class label). Each neuron processes the input variables and passes the values to the neuron in a subsequent layer. Each connection of neurons is assigned a certain weight '$w^{(4)}$' and bias '$\beta^{(4)}$'. The output of a hidden neuron is computed using the activation function '$\psi$' computed using 'M' neuron connections between the input and hidden layer in (6).

$$H_j = \psi(\beta^{(4)}_j + \sum_{k=1}^{M} W^{(4)}_{kj} X_j) \quad (6)$$

The classifier output is then computed by combining the 'L' outputs from the hidden nodes using a sigmoidal activation function '$\omega$' given by (7).

$$\Pi_j = \omega(\beta^{(4)}_j + \sum_{k=1}^{L} W^{(4)}_{kj} H_k) \quad (7)$$

During model estimation, the weights and biases of each neuron are initialized and iteratively varied to minimize classification error on the training data set. The number of neurons in the hidden layer can also be varied.

*2) Multi-class Classificaion Models*

The following classification models can be applied to data sets that contain three or more unique class labels. This category of classifiers also includes one-vs-all binary classification models.

*a) Logistic Regression*

Multinomial logit regression models use linear prediction function defined in (8).

$$\Lambda(i,j) = \beta^{(5)}_{0,i} + \beta^{(5)}_{1,i} X_{1,j} ... + \beta^{(5)}_{N^{(1)},i} X_{N^{(1)},j} = B_i X_j \quad (8)$$

The probability of a sample 'j' belonging to class label 'i', where 'C' is the number of unique classes is given in (9), where the parameter vectors $B_i$ are to be estimated.

$$\Pi^i_j = \frac{\exp(B_i X_j)}{\sum_{k=1}^{C} \exp(B_k X_j)} \quad (9)$$

*b) Neural Network*

The weight and bias vectors between neurons in the input layer, hidden layer and output layers defined in (6-7) can be trained to predict multiple class labels.

*c) Decision Forest*

Decision trees with nodes corresponding to feature values can be constructed with terminal leaf nodes corresponding to class labels. Each tree must then vote for the most popular lass label using. Thus, the binary classifier can be extended to accommodate multi-class decision making tasks.

*d) Decision Jungle*

These classifiers are an extension to decision forests and they are constructed using an ensemble of decision directed acyclic graphs. Decision jungles have two advantages over decision forests. First, they allow decision tree branches to merge, thereby resulting in decision acyclic graphs. This merging operation reduced memory utilization in spite of longer training time requirements. Second, decision jungles perform integrated feature selection and classification that minimize the effects of noisy features.

*e) One-vs-All Boosted Decision Tree, One-vs-All Support Vector Machine*

This category of classifiers trains binary classifiers to predict the samples of a particular class label versus all remaining class labels. Prediction is then performed by implementing the individual binary classifiers, and choosing the prediction with the highest probability.

*C. Feature Selection*

The MAMLS platform includes a Filter-Based Feature Reduction module. This module takes as input the training data set and outputs the label column and a variable number of top-ranked feature columns. These rankings are made according to Fisher Score, Mutual Information, or Chi Squared feature scoring methods. The ranked features are then sorted in ascending order of their scores and the top ranked scores represent the most discriminating features for classification. The feature ranking methods are described below.

*1) Fisher Score*

For binary classification tasks the mean and variance of feature 'l' computed using samples from class label 0 and 1 are $(\mu_l^0, \mu_l^1)$ and $(\sigma_l^{02}, \sigma_l^{12})$, respectively. The fisher score of each of the 'd'-features is computed using (10).

$$V(l) = \frac{(\mu_l^1 - \mu_l^0)^2}{(\sigma_l^{02} + \sigma_l^{12})} \quad (10)$$

*2) Mutual Information*

The importance of each feature can be assessed in terms of mutual information computed using (11).

$$I(l) = \sum_{x_l=x_1}^{x_d} \sum_{y=0}^{i} \log \frac{P(X=x_l, Y=i)}{P(X=x_l)P(Y=i)} \quad (11)$$

Here, $P(X=x_l, Y=i)$ represents the proportion of samples with feature value $x_l$ and class label 'i'. Variations to the mutual information-based feature ranking include minimal-redundancy-maximal-relevance (mRMR) criterion [12].

*3) Chi-Squared*

For binary classification tasks, the $\chi^2$ statistic for each feature is defined using (12).

$$\chi^2(l) = \frac{N^{(1)}}{P(X=x_l).P(X \neq x_l).P(Y=1).P(Y=0)} * \qquad (12)$$
$$[P(X=x_l,Y=1).P(X \neq x_l,Y=0) - P(X=x_l,Y=0).P(X \neq x_l,Y=1)]^2$$

## III. EXPERIMENTS AND RESULTS

*A. Experiments*

Two sets of experiments are performed. First, the MAMLS platform is benchmarked on three public data sets. Second, the Generalized Flow is evaluated on the locally generated multi-class data set for DR lesion classification. Further, the performance of multi-class classifiers is compared to hierarchical binary classification accuracy.

*1) Benchmarking MAMLS platform*

The performance of classification models is benchmarked for three public data sets. First, the models are optimally parameterized by 5-fold cross validation [7]. Then, dimensionality reduction is performed by choosing only the highest raked features. The classifier parameters and most discriminating set of ordered feature set that maximizes classification accuracy on the training data set are chosen as optimal.

*a) Optimal Parameterization*

For this step, the training data set is partitioned into 5-folds and then input to a "Sweep Parameters" module. This module is designed to determine the optimal parameters for each model from the 5-folds using accuracy as the metric for maximizing each classification model's performance. This module output, a trained, parameterized classification model, is tested using the testing data set within a scoring module and the classification performance metrics are then evaluated. The result of this process is a Receiver Operating Characteristic (ROC) curve for binary classification data sets and a confusion matrix for multi-class classification data sets.

*b) Dimensionality Reduction*

For each public data set, each optimally parameterized classification model is re-trained, tested, and evaluated using the reduced feature data set; the output of the Filter-Based Feature Selection module. As the number of top ranked features used for classification is varied as $l = [1, 2 ....d]$, the feature set that maximizes the classification accuracy is selected as the optimal most discriminating feature set.

*2) Generalized Flow*

The designed Generalized Flow accepts a single data set input and outputs the results of the best performing classification model with the optimal feature set. It employs decision-making scripts in 'R', optimal classification model selection and its parameterization to maximize classification accuracies. Classification models are scored based on the average validation error on the training data set. So far, no two models have been scored similarly in any of the evaluated data sets. For future data sets, classification model complexity will be used as a metric for model selection in case two models are scored similarly. Fig. 2 illustrates the steps of the proposed method that combines the built-in MAMLS modules with the new 'R'-based decision making modules.

In Fig 2, Decision 1 directs the input data set to either the binary or multi-class classification groups. Within these classification groups, Decision 2 trains, tests, and evaluates the data set using the classification models described in Section II B and selects the highest performing model. After the multi-class classification group, Decision 3 is used to decide if hierarchical binary classification would enhance average multi-class classification accuracy. Finally, the maximum classification metrics, thus obtained, are returned to the user.

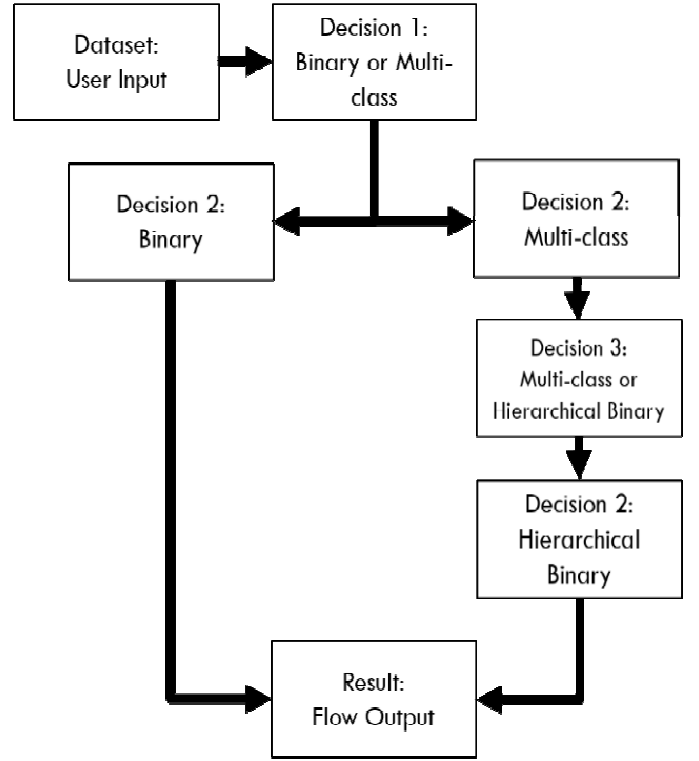

Fig 2. Block diagram of the proposed Generalized Flow.

*B. Results*

The performance of some of the existing state-of-the-art methods are compared to the proposed Generalized Flow outputs. For binary classification tasks, the number of samples belonging to class 1 also classified as class 1 (true positives, tp), the number of sample belonging to class 0 also classified as class 0 (true negatives, tn), the number of samples belonging to class 1 misclassified as class 0 (false negatives, fn) and the number of samples belonging to class 0 misclassified as class 1 (false positives, fp) on the testing data set are evaluated. The metrics for assessing the classification performance for binary classification tasks is given in (13).

$$precision^{(i=1)} = \frac{tp^{(i=1)}}{(tp^{(i=1)} + fp^{(i=1)})}, recall^{(i=1)} = \frac{tp^{(i=1)}}{(tp^{(i=1)} + fn^{(i=1)})},$$

$$accuracy^{(i=1)} = \frac{tp^{(i=1)} + tn^{(i=1)}}{N^{(2)}} \quad (13)$$

Here, the total number of samples in the testing data set is $N^{(2)} = 0.7N$. For multi-class classification tasks, the precision, recall and accuracy for each class vs the samples from all other classes is micro-averaged and macro-averaged. While micro-averaging represents the weighted average based on the frequency of samples from each class, macro-averaging represents unweighted mean of precision, recall and accuracy metrics [13]. If the total number of classes is 'C', then for each class $i = \{0,1,..C\}$, the number of samples in each class varies as $n^{(0)}, n^{(1)}...n^{(C)}, \sum_{i=0}^{C} n^{(C)} = N^{(2)}$. Then the micro-averaged metrics are computed using (14) and the macro-averaged metrics are computed using (15).

$$precision = \frac{1}{N^{(2)}} \sum_{i=0}^{C} n^{(i)} precision^{(i)}, recall = \frac{1}{N^{(2)}} \sum_{i=0}^{C} n^{(i)} recall^{(i)},$$

$$accuracy = \frac{1}{N^{(2)}} \sum_{i=0}^{C} n^{(i)} accuracy^{(i)} \quad (14)$$

$$precision = \frac{1}{C} \sum_{i=0}^{C} precision^{(i)}, recall = \frac{1}{C} \sum_{i=0}^{C} recall^{(i)},$$

$$accuracy = \frac{1}{C} \sum_{i=0}^{C} accuracy^{(i)} \quad (15)$$

Both micro and macro-averaged metrics are important since micro-averaged metrics tend to weigh the most frequent class heavily while macro-averaging considers all classes to be equally significant.

The impact of dimensionality reduction by top feature set selection, comparison of classification performance with prior work and the optimally parametrized classification metrics obtained on the public and local data sets are given below.

*1) Impact of Dimensionality Reduction*

The effect of dimensionality reduction is analyzed by incrementally varying the number of top ranked features used for classification from one feature to the all the available features. The trend of classification accuracy by varying the number of classifying features used for binary classification is observed for the Wisconsin Breast Cancer [2] data set in Fig. 3. Fig. 4 shows the variation in accuracy versus the number of features for the German Credit Card [4] data set and Fig. 5 shows the trends for the Telescope data set [6]. These figures illustrate that there exists a robust subset of features that maximizes classification accuracy regardless of the total number of features contained in the data set.

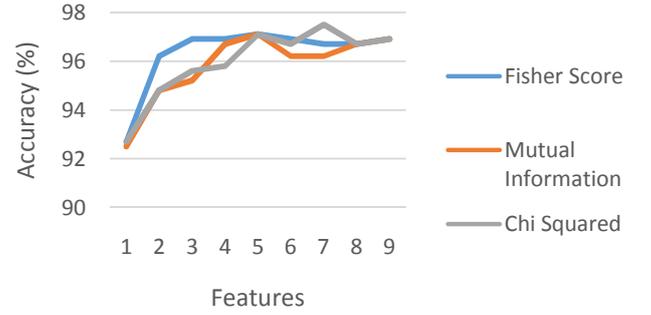

Fig. 3. Wisconsin Breast Cancer [2] dimensionality reducsion results with a maximum at top 7 features (Chi Squared) followed by Fisher-scoring (10) using top 5 features. Thus, maximum accuracyoccurs by using 7 features ranked by Chi Squared statistic (12).

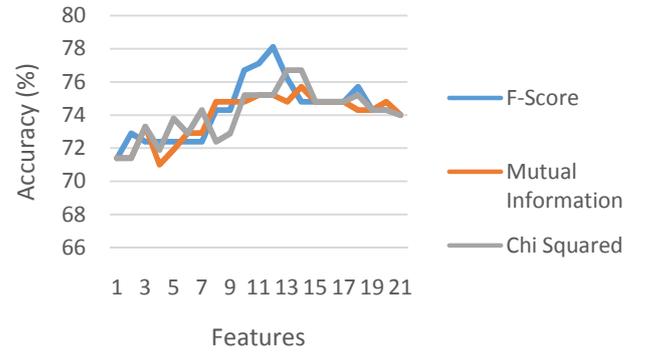

Fig. 4. German Credit Card [4] dimensionality reduction results with a maximum accuracy at top 12 features using Fisher Score (10).

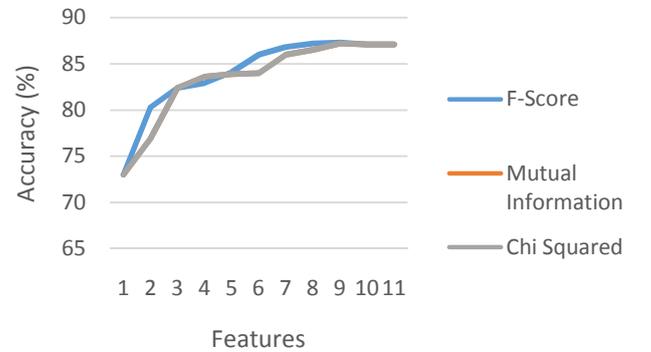

Fig. 5. Telescope [6] dimensionality reduction results with a maximum accuracy at top 9 features using the Fisher Score statistic (10). In this figure, the Mutual Information (11) trend coincides with that of the Chi Squared (12).

*2) Comparison with Prior Work*

Table I shows the comparison between some existing state-of-the-art methods and the proposed MAMLS Generalized Flow in terms of binary classification accuracy. Here, we observe that the classification accuracy using the proposed flow increases for all but the German Credit Card data set. However, on the German Credit card data set, the proposed flow does not discard any data samples as compared to the method in [5], since the proposed flow disregards features with incomplete values (as they get ranked low). Hence the proposed flow is capable of classifying more data samples by automatically disregarding incomplete features.

TABLE I. COMPARISON OF THE PROPOSED FLOW WITH EXISTING METHODS

| Data Set | $accuracy^{(i=1)}$ (%) | |
|---|---|---|
| | **Existing Method** | **Proposed flow** |
| Wisconsin Breast Cancer | [3] 96.6 | 97.5 |
| German Credit Card | [5] 83.3 | 78.1 |
| Telescope | [8] 86.94 | 87.3 |
| DIARETDB1 – Hierarchiy level 2 | [9] 89 | 95.7 |
| DIARETDB1 – Hierarchy level 3 | [9] 80 | 85.3 |

Also, in Table I, we observe that the existing method in [9] is used for screening of images with DR, and hence it is fine tuned to maximize classification recall (to ensure no patient with DR is missed) at the cost of low precision. However, the proposed flow is directed to increase the overall classification accuracy of bright and red lesions.

*3) Classification of Public Data Sets*

For each public data set, the benchmarked MAMLS platform is analyzed in terms of the optimally parameterized classification metrics achieved using the proposed flow.

Table II shows the best performing classification models and their results for each public data set. The highest performing classifier and its parameter settings are also presented in Table II.

TABLE II. CLASSIFICATION PERFORMANCE ON PUBLIC DATA SETS. AUC IS THE AREA UNDER ROC CURVES

| Data Set | Classification Model | Optimal Parameters | $accuracy^{(i=1)}$ (%) | $precision^{(i=1)}$ | $recall^{(i=1)}$ | AUC |
|---|---|---|---|---|---|---|
| Wisconsin Breast Cancer | Binary SVM | Top 6 Features: F-Score | 99.7 | 0.687 | 0.758 | 0.997 |
| | | Lambda = 1E-06 | | | | |
| German Credit Card | Binary Neural Network | Top 5 Features: Mutual Information | 77.1 | 0.693 | 0.615 | 0.83 |
| | | Learning Rate = 0.04 | | | | |
| | | Number of Hidden Nodes =100 | | | | |
| Telescope | Binary Boosted Decision Tree | Top 9 Features: F-Score | 87.3 | 0.85 | 0.776 | 0.922 |
| | | Number of Leaves = 20 | | | | |
| | | Learning Rate = 0.2 | | | | |
| | | Number of Trees = 200 | | | | |

*4) Classification of Local DIARETDB1 Data Set*

The proposed Generalized Flow is capable of receiving multi-class and binary input data and processing them separately. Since the local data set derived from the DIARETDB1 data is a multi-class classification task, we analyze the performance of multi-class classifiers and hierarchical binary classifiers on this data set to maximize the micro and macro-averaged classification accuracies.

*a) DIARETDB1 Multi-class Classification*

The multi-class classification performance on the local data set with is represented in Fig. 6 in terms of a confusion matrix taken from MAMLS experiment canvas. In Fig. 6 we observe that the uneven distribution of class labels within this data set contribute to a poor positive classification for class labels 1, 4, and 5. The micro and macro-averaged classification metrics thus achieved for the multi-class classification task are shown in Table III. To further improve the overall classification accuracy on this multi-class data set, we analyze the best achievable macro-averaged recall of multi-class classification with respect to randomized classification.

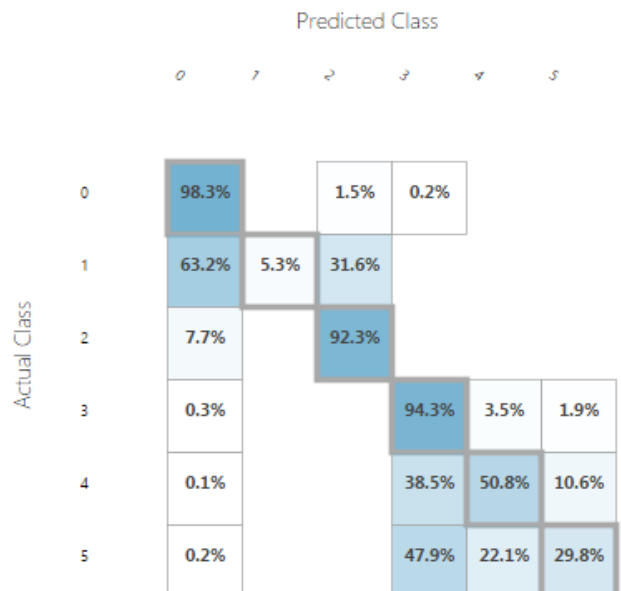

Fig. 6. Multi-class classification on DIARETDB1 [10] local data set using one-vs-all boosted decision tree classifier.

Randomized classification is the instance when all the samples are classified as the class label with the highest frequency. The goal here is to design a classifier group (group of hierarchical classifiers) that has better macro-averaged recall when compared to the recall of the randomized classifier. In this case, the DIARETDB1 [10] contains 15,945 total samples with 10,967 belonging to class label 3 (class 3 has highest frequency). If all the samples are classified as class 3, the randomized recall is 68.78%. From Table III, we observe that the macro-averaged recall of best multi-class classification is 61.8%, which is lower than randomized recall. Thus, there must be a group of hierarchical classifiers that can improve macro-averaged recall more than the multi-class classifiers and the randomized classification.

The hierarchical set of binary classifiers are implemented using the hierarchies defined in Section IIA4.

The binary classification metrics in each hierarchy level is shown in Table IV. We observe that for hierarchical classification, the macro-averaged accuracy, precision, and recall by combining the classifiation metrics from all the hierarchy levels are 89.72%, 88%, and 82.3%, respectively. Thus, the macro-averaged recall by hierarchical classification improves the recall when compared to multi-class and reandomized classification.

In Table IV we observe that the hierarchy level 2 has [precision, recall, AUC] of [0.975, 0.905, 0.976], which is a significant improvement over the metrics [0.647,0.944,0.959] achieved by the method in [9]. Also, for the hierarchy level 3, the proposed flow achieves [precision, recall, AUC] of [0.727, 0.656, 0.89], which is a significant improvement over the metrics [0.332,0.755,0.866] achieved by the method in [9].

TABLE III. MULTI-CLASS CLASSIFICATION METRICS ON DIARETDB1 [10] LOCAL DATA SET.

| Features | Classification Model | Optimal Parameterization | Accuracy (%) | | Precision | | Recall | |
|---|---|---|---|---|---|---|---|---|
| | | | Micro | Macro | Micro-averaged | Macro-averaged | Micro-averaged | Macro-averaged |
| All Features | One-vs-All Boosted Decision Tree | Number of Leaves = 20 | 82.1521 | 94.0507 | 0.8215 | 0.8053 | 0.821521 | 0.618066 |
| | | Learning Rate = 0.1 | | | | | | |
| | | Number of Trees = 181 | | | | | | |

TABLE IV. HIERARCHICAL BINARY CLASSIFICAION ON DIARETDB1[10] LOCAL DATA SET.

| Hierarchical Level | Features | Best Classification Model | Optimal Parameterization | Accuracy (%) | Precision | Recall | AUC |
|---|---|---|---|---|---|---|---|
| Hierarchy level 1 | Top 10 featuresselected using mRMR [7] | Binary Boosted Decision Tree | Number of Leaves - 20 | 99.4 | 0.995 | 0.998 | 0.993 |
| | | | Learning Rate - 0.1 | | | | |
| | | | Number of Trees - 67 | | | | |
| Hierarchy level 2 | All Features | Binary Decision Forest | Random Split Count - 128 | 95.7 | 0.975 | 0.905 | 0.976 |
| | | | Maximum Depth - 64 | | | | |
| | | | Ensemble Element Count - 8 | | | | |
| Hierarchy level 3 | All Features | Binary Boosted Decision Tree | Number of Leaves - 20 | 85.3 | 0.727 | 0.656 | 0.89 |
| | | | Learning Rate - 0.1 | | | | |
| | | | Number of Trees - 200 | | | | |
| Hierarchy level 4 | All Features | Binary Support Vector Machine | Lambda - 1 E-06 | 97.1 | 0.976 | 0.993 | 0.962 |
| Hierarchy level 5 | All Features | Binary Decision Forest | Random Split Count - 1024 | 71.7 | 0.727 | 0.562 | 0.775 |
| | | | Maximum Depth - 64 | | | | |
| | | | Ensemble Element Count - 32 | | | | |

Finally, in Fig. 7, the ROC curves obtained for each binary hierarchy classification level in the local DIARETDB1 data set is shown. The MAMLS experiment canvas at the end of the evaluate module returns these ROC curves to the end user. The classification threshold values can be manually changed to obtain the best precision/recall metric for that particular classification step. In Fig. 7 we observe that as the hierarchical levels increase from level 1 through level 5, the AUC reduces. This observation is intuitive since classification of bright lesions from red lesions (hierarchy level 1 task) is easier than classification of hemorrhages from micro-aneurysms (hierarchy level 5 task). Additionally, the classification metrics obtained in each hierarchy level is significantly better than the existing method in [9]. These observations motivate the use of the proposed Generalized Flow for data sets derived from medical images. As the size of the image data sets grow, the proposed cloud-based flow will automatically parameterize to return the maximized overall macro-averaged precision, recall, accuracy and AUC.

## IV. CONCLUSIONS

The MAMLS platform provides classification modules that can be fine-tuned and combined with dimensionality reduction and decision-making modules for classification tasks on scalable "Big Data" sets. In this work, we demonstrate and benchmark the performance of a novel Generalized Flow in the MAMLS platform that maximizes overall classification accuracies independent of computation device limitations. All experiments are conducted on a laptop with Intel i3 (2.5 GHz) and 6 GB RAM accessing the MAMLS platform. The timings for the Wisconsin Breast Cancer, Telescope, and DIARETDB1 data sets in minutes are 99, 96, and 35, respectively. These timings can be evaluated using the starting-time and the ending-time time stamps in the MAMLS experiment canvas. Future efforts will be directed towards benchmarking more public data sets on the MAMLS platform and analyzing additional medical images such as Computed Tomography (CT images) and MRI images using the proposed flow. Additionally, regression tasks can also be integrated with the existing classification modules in future works.

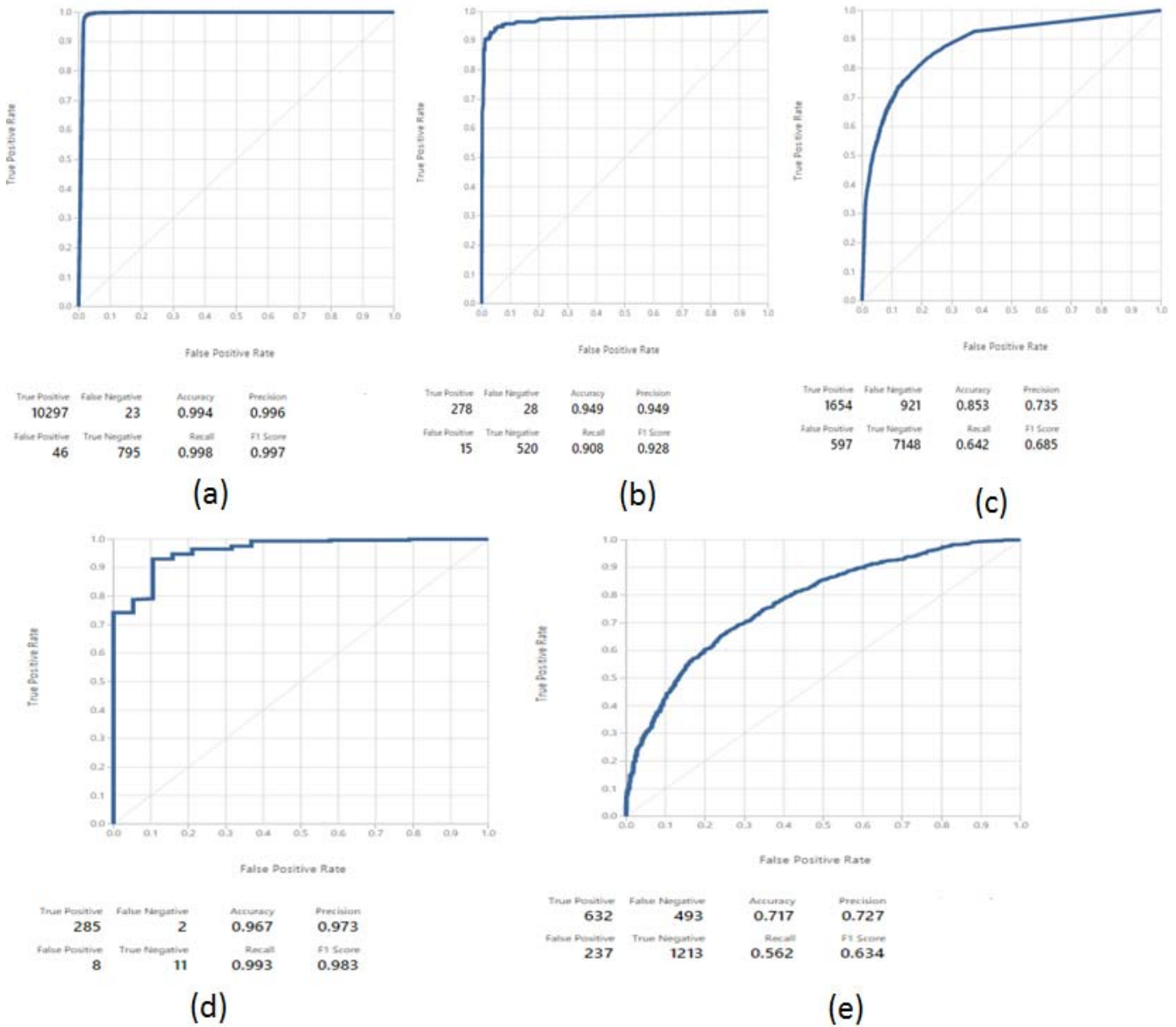

Fig. 7. DIARETDB1 hierarchical classification ROC curves. (a) level 1. (b) level 2. (c) level 3. (d) level 4. (e) level 5.

V. ACKNOWLEDGEMENTS

This work is supported in parts by Washington NASA Space Grant Consortium, University of Washington Bothell, the Office of Research, and Microsoft Azure ML Research Award.

REFERENCES

[1] Chen, Min, et al. Big data: related technologies, challenges and future prospects. Springer, 2014.

[2] M. Lichman (2013). UCI Machine Learning Repository [https://archive.ics.uci.edu/ml/datasets/Breast+Cancer+Wisconsin+%28Original%29]. Irvine, CA: University of California, School of Information and Computer Science.

[3] Akay, Mehmet Fatih. "Support vector machines combined with feature selection for breast cancer diagnosis." Expert systems with applications 36.2 (2009): 3240-3247.

[4] M. Lichman (2013). UCI Machine Learning Repository [https://archive.ics.uci.edu/ml/datasets/Statlog+(German+Credit+Data)]. Irvine, CA: University of California, School of Information and Computer Science.

[5] Huang, Cheng-Lung, Mu-Chen Chen, and Chieh-Jen Wang. "Credit scoring with a data mining approach based on support vector machines." Expert systems with applications 33.4 (2007): 847-856.

[6] M. Lichman (2013). UCI Machine Learning Repository [https://archive.ics.uci.edu/ml/datasets/MAGIC+Gamma+Telescope]. Irvine, CA: University of California, School of Information and Computer Science.

[7] Cherkassky, Vladimir, and Filip M. Mulier. Learning from data: concepts, theory, and methods. John Wiley & Sons, 2007.

[8] Kou, Gang, et al. "Evaluation of classification algorithms using MCDM and rank correlation." International Journal of Information Technology & Decision Making 11.01 (2012): 197-225.

[9] S. Roychowdhury, D. Koozekanani, and K. Parhi, "Dream: Diabetic retinopathy analysis using machine learning," IEEE Journal of Biomedical and Health Informatics, vol. 18, no. 5, pp. 1717–1728, Sept 2014.

[10] Kauppi, Tomi, et al. "The DIARETDB1 Diabetic Retinopathy Database and Evaluation Protocol." BMVC. 2007.

[11] S. Roychowdhury, Sohini, K. Parhi, and D. Koozekanani. "Method and apparatus to detect lesions of diabetic retinopathy in fundus images." U.S. Patent Application 14/120,027

[12] Peng, Hanchuan, Fuhui Long, and Chris Ding. "Feature selection based on mutual information criteria of max-dependency, max-relevance, and min-redundancy." Pattern Analysis and Machine Intelligence, IEEE Transactions on 27.8 (2005): 1226-1238.

[13] Yang, Yiming. "An evaluation of statistical approaches to text categorization." Information retrieval 1.1-2 (1999): 69-90.